\documentclass[9pt,arxiv]{lapreprint}

\usepackage[version=4]{mhchem} 
\usepackage{siunitx}    
\usepackage{pdflscape}  
\usepackage{rotating}   
\usepackage{textgreek}  
\usepackage{gensymb}    
\usepackage[misc]{ifsym} 
\usepackage{orcidlink}  
\usepackage{listings}   
\usepackage{colortbl}   
\usepackage{tabularx}   
\usepackage{longtable}  
\usepackage{subcaption}
\usepackage{multirow}
\usepackage{snotez}     
\usepackage{csquotes}   
\usepackage{annotate-equations}
\usepackage{tabularx} 
\usepackage{booktabs}

\DeclareSIUnit\Molar{M}

\usepackage[
    backend=biber,
    style=nature,       
    sorting=none,       
    natbib=true,
    hyperref=true,
    alldates=year,
    uniquename=false,
    maxbibnames=99,     
]{biblatex}

\AtEveryBibitem{
    \clearfield{urlyear}
    \clearfield{urlmonth}
    \clearlist{language}
}

\title{Image-based Quantification of Postural Deviations on Patients with Cervical Dystonia: A Machine Learning Approach Using Synthetic Training Data}
\author[ \orcidlink{0000-0002-7590-7286} 1]{Roland Stenger}
\author[ \orcidlink{0000-0003-3067-5239} 2,3]{Sebastian Löns}
\author[ \orcidlink{0009-0006-2039-423X} 4]{Nele Brügge}
\author[ \orcidlink{0000-0002-1324-2776} 3]{Feline Hamami}
\author[ \orcidlink{0000-0002-3219-2284} 3]{Alexander Münchau}
\author[ \orcidlink{0009-0009-8565-6060} 3,5]{Theresa Paulus}
\author[ \orcidlink{0000-0002-3569-6242} 3]{Anne Weissbach}
\author[ 3]{Gesine M. Sallandt}
\author[ \orcidlink{0000-0001-7335-8888} 6]{Tatiana Usnich}
\author[ \orcidlink{0000-0002-9651-5986} 5,6]{Max Borsche}
\author[ \orcidlink{0000-0002-7794-0282} 5,6]{Martje G. Pauly}
\author[ \orcidlink{0000-0002-7162-9821} 5,6]{Lara M. Lange}
\author[ \orcidlink{0000-0002-7558-8107} 5,7]{Markus A. Hobert}
\author[ \orcidlink{0000-0002-5906-5947} 5]{Rebecca Herzog}
\author[ \orcidlink{0000-0002-3291-7222} 8]{Ana Luísa de Almeida Marcelino}
\author[ \orcidlink{0000-0002-0597-2648} 8]{Tina Mainka}
\author[ \orcidlink{0000-0001-9650-6820} 8,9]{Friederike Schumann}
\author[ \orcidlink{0000-0001-6026-185X} 8]{Lukas L. Goede}
\author[ \orcidlink{0000-0002-9524-7836} 8]{Johanna Reimer}
\author[ 7]{Kirsten E. Zeuner}
\author[ \orcidlink{0009-0004-3539-9362} 7]{Julienne Haas}
\author[ \orcidlink{0000-0002-3532-3155} 7]{Jos Becktepe}
\author[ \orcidlink{0000-0002-8035-6271} 7]{Alexander Baumann}
\author[ \orcidlink{0009-0000-8554-360X} 7]{Robin Wolke}
\author[ \orcidlink{0000-0003-0484-385X} 10]{Chi Wang Ip}
\author[ \orcidlink{0000-0001-9042-7689} 10]{Thorsten Odorfer}
\author[ \orcidlink{0000-0003-3393-5657} 10]{Daniel Zeller}
\author[ \orcidlink{0000-0002-6927-9105} 10]{Lisa Harder-Rauschenberger}
\author[ \orcidlink{0000-0001-9622-0216} 11]{John-Ih Lee}
\author[ \orcidlink{0000-0001-7987-658X} 11,12]{Philipp Albrecht}
\author[ 11]{Petyo Nikolov}
\author[ \orcidlink{0009-0002-4075-669X} 11]{Tristan Kölsche}
\author[ \orcidlink{0000-0002-0675-9064} 13]{Joachim K. Krauss}
\author[ \orcidlink{0009-0007-5082-0460} 13]{Johanna M. Nagel}
\author[ \orcidlink{0009-0009-9319-4724} 13]{Joachim Runge}
\author[ 13]{Jessica Utermarck}
\author[ 14]{Katja Kollewe}
\author[ \orcidlink{0009-0009-4722-9914} 15]{Johanna Doll-Lee}
\author[ 15]{Johanne Heine}
\author[ 15]{Linda Veith Sanches}
\author[ \orcidlink{0000-0002-3767-6376} 16]{Simone Zittel}
\author[ \orcidlink{0000-0003-3561-3328} 16]{Kai Grimm}
\author[ \orcidlink{0000-0002-4234-5635} 17]{Pawel Tacik}
\author[ \orcidlink{0000-0002-5557-3127} 18,19]{André Lee}
\author[ \orcidlink{0000-0003-1014-1782} 3]{Tobias Bäumer}
\author[ \orcidlink{0000-0002-3553-5131} 2,20 \Letter]{Sebastian Fudickar}

\affil[1]{Institute of Medical Informatics, University of Lübeck, Lübeck, Germany}
\affil[2]{Lübeck Centre for Rare Diseases, University Medical Center Schleswig-Holstein, Lübeck, Germany}
\affil[3]{Institute of Systems Motor Science, University of Lübeck, CBBM, Lübeck, Germany}
\affil[4]{German Research Center for Artificial Intelligence, AI in Medical Image and Signal Processing, Lübeck, Germany}
\affil[5]{Department of Neurology, University Medical Center Schleswig-Holstein, Campus Lübeck, Lübeck, Germany}
\affil[6]{Institute of Neurogenetics, University of Lübeck, Lübeck, Germany}
\affil[7]{Department of Neurology, University Medical Center Schleswig-Holstein, Campus Kiel, Kiel, Germany}
\affil[8]{Department of Neurology with Experimental Neurology, Movement Disorders and Neuromodulation Unit, Charité-Universitätsmedizin Berlin, Corporate member of Freie Universität Berlin, Humboldt-Universität zu Berlin, and Berlin Institute of Health (BIH), Berlin, Germany}
\affil[9]{Department of Neurology, Park-Klinik Weißensee, Berlin, Germany}
\affil[10]{Department of Neurology, University Hospital Würzburg, Würzburg, Germany}
\affil[11]{Department of Neurology, Medical Faculty and University Hospital Düsseldorf, Heinrich-Heine-University Düsseldorf, Düsseldorf, Germany}
\affil[12]{Department of Neurology, Maria Hilf Clinics, Mönchengladbach, Germany}
\affil[13]{Department of Neurosurgery, Hannover Medical School, Hannover, Germany}
\affil[14]{Department of Neurology, International Neuroscience Institute Hannover, Hannover, Germany}
\affil[15]{Department of Neurology, Hannover Medical School, Hannover, Germany}
\affil[16]{Department of Neurology, University Medical Center Hamburg-Eppendorf, Hamburg, Germany}
\affil[17]{Department of Parkinson’s disease, Sleep and Movement Disorders, University Hospital Bonn, Bonn, Germany}
\affil[18]{Institute of Music Physiology and Musicians’ Medicine, University of Music, Drama and Media Hannover, Germany}
\affil[19]{Department of Neurology, TUM Klinikum Rechts der Isar, Munich, Germany}
\affil[20]{Section for Clinical Research IT (SKFIT), Institute of Medical Biometry and Statistics (IMBS), University of Lübeck, Lübeck, Germany}

\metadata[]{\Letter\hspace{.5ex} For correspondence}{sebastian.fudickar@uni-luebeck.de}
\metadata[]{Author information}{These authors contributed equally: Roland Stenger, Sebastian Loens | These authors share last authorship: Tobias Bäumer, Sebastian Fudickar}
\metadata[]{Ethics declaration}{Images of real CD patients were extracted from the DysTract Dystonia registry (https://www.isms.uni-luebeck.de/en/research/dystract), a multicenter cross-sectional study of adults with dystonia. All procedures were performed in compliance with relevant laws and institutional guidelines. The study was approved by the local ethics board (ethics vote no. 16-197). All patients gave written informed consent for participation and the use of their imaging data. The privacy rights of all human subjects were strictly observed throughout the study.}
\metadata[]{Acknowledgements}{The authors thank all participants of the study for their time and effort, reviewing dystonia patient images. We further thank the patients depicted for their consent.}
\metadata[]{Competing interests}{The authors declare no competing interests.}

\leadauthor{Roland Stenger}
\shorttitle{Image-based Quantification of Postural Deviations on Patients with Cervical Dystonia}

\begin{document}
\maketitle
\begin{abstract}

Cervical dystonia (CD) is the most common form of dystonia, yet current assessment relies on subjective clinical rating scales, such as the Toronto Western Spasmodic Torticollis Rating Scale (TWSTRS), which requires expertise, is subjective and faces low inter-rater reliability some items of the score. To address the lack of established objective tools for monitoring disease severity and treatment response, this study validates an automated image-based head pose and shift estimation system for patients with CD. We developed an assessment tool that combines a pretrained head-pose estimation algorithm for rotational symptoms with a deep learning model trained exclusively on ~16,000 synthetic avatar images to evaluate rare translational symptoms, specifically lateral shift. This synthetic data approach overcomes the scarcity of clinical training examples. The system's performance was validated in a multicenter study by comparing its predicted scores against the consensus ratings of 20 clinical experts using a dataset of 100 real patient images and 100 labeled synthetic avatars. The automated system demonstrated strong agreement with expert clinical ratings for rotational symptoms, achieving high correlations for torticollis (r=0.91), laterocollis (r=0.81), and antero-/retrocollis (r=0.78). For lateral shift, the tool achieved a moderate correlation (r=0.55) with clinical ratings and demonstrated higher accuracy than human raters in controlled benchmark tests on avatars. By leveraging synthetic training data to bridge the clinical data gap, this model successfully generalizes to real-world patients, providing a validated, objective tool for CD postural assessment that can enable standardized clinical decision-making and trial evaluation.

\end{abstract}
\section{Introduction} \label{intro}
Dystonia is a movement disorder characterized by sustained or intermittent abnormal movements, postures, or both \cite{albaneseDefinitionClassificationDystonia2025}. The etiology remains unknown for most patients except for acquired or monogenic cases that make up less than 5\% \cite{zechMonogenicVariantsDystonia2020}, and in these idiopathic cases, no biomarkers or imaging markers are available yet \cite{smitDystoniaManagementWhat2021}. The most common form of dystonia is cervical dystonia (CD), characterized by intermittent or sustained postures and/or tremor of the head and neck \cite{albaneseDefinitionClassificationDystonia2025,velickovicCervicalDystoniaPathophysiology2001,albaneseIsolatedCervicalDystonia2023}. Diagnosis is based on clinical evaluation, and the severity of symptoms is typically quantified using clinical rating scales, which are at least partially subjective. The Toronto Western Spasmodic Torticollis Rating Scale (TWSTRS) is the most widely used rating scale of its kind \cite{comellaTeachingTapeMotor1997,jostRatingScalesCervical2013,chrobakContentOverlap912024,wollnerSubjectiveObjectiveSymptom2021}. Treatment options of CD include injection of botulinum toxin (BoNT) in hyperactive muscles, which is the gold standard treatment to date \cite{rodriguesBotulinumToxinType2020,martinez-nunezAdjuvantMedicalTherapy2022}, and also physical therapy and deep brain stimulation \cite{loudovici-krugPhysiotherapyCervicalDystonia2022,kassayeEffectivenessPhysiotherapyPatients2024,bledsoeTreatmentDystoniaMedications2020}. To monitor severity of CD and treatment response, accurate clinical assessment is needed. It can also directly influence clinical decision-making regarding BoNT treatments, e.g., the dose and location of treatment \cite{castagnaManagementCervicalDystonia2019,albanesePracticalGuidanceCD2015,swopeTreatmentRecommendationsPractical2008}. For this reason, accurate and consistent clinical assessment of CD is critical for monitoring disease progression, evaluating treatment response.
Currently, no clinically established tools exist to objectively diagnose or monitor the course of CD. While numerous objective biomarkers and diagnostic approaches have been proposed, like neurofilament-light levels, ultrasound, serum metabolomic, or plasma proteomic panels \cite{castagnaManagementCervicalDystonia2019}, \cite{timsinaBloodBasedProteomicsAdultOnset2024,liuMetabolomicStudyCervical2021,gelisinEvaluationMiR526b3p2023,ferrazzanoNeurofilamentAssessmentPatients2022,rafeeWearableDeviceMeasure2023,trinidad-fernandezHeadMountedDisplayClinical2023,berbakovQuantitativeAssessmentHead2019,loramObjectiveAnalysisNeck2020,fietzekRoleUltrasoundPersonalized2021,mullerHomocysteineSerumMarkers2005,scorrExplorationPotentialImmune2024,valerianiMicrostructuralNeuralNetwork2020}, none of these methods have shown sufficient robustness or scalability for routine clinical use. Each is limited by practical barriers, such as high complexity, the need for specialized equipment, invasiveness, or sensitivity to environmental conditions, which collectively hinder their integration into everyday clinical practice. Therefore, no broadly validated objective biomarker is available for everyday CD evaluation and treatment decisions, which continues to rely on expert clinical judgment.

Computer Vision has been gaining attention as a tool for clinical evaluation of movement disorders in recent years \cite{yangFastEvalParkinsonismInstant2024,bruggeAutomatedMotorTic2023,pecoraroComputerVisionTechnologies2025,parkArtificialIntelligencebasedVideo2024}. Several studies have demonstrated that (depth) image- or video-based computer vision methods can support symptom rating in the context of CD \cite{yePilotFeasibilityStudy2022,peachHeadMovementDynamics2024,nakamuraPilotFeasibilityStudy2019,zhangHoldThatPose2022}. All these studies aim to estimate individual TWSTRS item scores directly or relate angular head pose estimations to TWSTRS items. All approaches rely on keypoint extraction using methods such as Azure Kinect body tracking, Mediapipe facemesh \cite{MediaPipeFrameworkPerceiving}, YOLOv3 \cite{redmonYOLOv3IncrementalImprovement,redmonYouOnlyLook2016}, or OpenFace 2.0 \cite{baltrusaitisOpenFace20Facial2018}. While these keypoints simplify the visual complexity of patient images into a low-dimensional representation, this comes at the cost of information. However, they allow for the manual definition of rules to map keypoints to TWSTRS items. These rule-based methods have the advantage of not requiring training data and provide transparent, explainable estimations. They have demonstrated to yield reasonable estimations for the rotational symptoms torticollis, laterocollis, and anterocollis/retrocollis, with correlations between the algorithmic estimations and ratings by clinical experts between 0.67 and 0.902, depending on the item \cite{yePilotFeasibilityStudy2022,peachHeadMovementDynamics2024,nakamuraPilotFeasibilityStudy2019,zhangHoldThatPose2022}. 

However, these approaches face two limitations: First, the lack of reliable ground truth makes objective performance evaluation difficult, as results are assessed only through correlation with potentially inconsistent human ratings. This leaves systematic errors undetected, either from the algorithm or the annotators. Second, less common symptoms such as lateral and sagittal shift are rarely addressed. Inter-rater reliability for these items is typically low \cite{jostRatingScalesCervical2013,kassayeInterdisciplinaryConsensusEvaluating2024}, and existing datasets contain very few positive cases, making evaluation unreliable. For example, Nakamura et al. \cite{nakamuraPilotFeasibilityStudy2019} report the only shift estimation to date, but base their evaluation on just 30 subjects, of whom only two showed shift symptoms.

To overcome these limitations, we adopted a data synthesis approach that enables systematic algorithm development and evaluation even in the absence of reliable human labels. Synthesized images provide precise and noise-free ground truth derived directly from the avatar rig, thereby disentangling algorithmic error from human annotation uncertainty at evaluation. This enables us not only to train models, but also to validate them to an extent that would not be possible on purely human-annotated datasets. For the lateral shift item, we train a regression model on ~16,000 rendered avatar images in total, representing a broad spectrum of lateral shifts. Rotational items are estimated via head pose estimation from a single frontal image using a pretrained neural network \cite{hempel6dRotationRepresentation2022}, with angular outputs mapped to TWSTRS scores.
The evaluation aims to address two research questions: First, how accurately does the algorithm estimate TWSTRS items on synthetic avatars compared to human raters? This benchmark against ground truth estimates how accurately the algorithm estimates TWSTRS items on synthetic avatars in comparison to human raters. Second, how well does the algorithm reproduce human expert ratings on real patient images, where averaged clinical annotations serve as a reference. Answering the first question enables an assessment of algorithmic accuracy in a controlled setting, while the second assesses the algorithm's relationship to the current, subjective clinical standard for clinical validity.

The main contributions of this work are:
Algorithmic Pipeline: We propose a method using head-pose estimation to quantify rotations (torti-, latero-, and antero/retrocollis) and a deep-learning model to estimate the lateral shift.
Open Dataset: We publicly release a CD-focused dataset containing about 16,000 labeled synthetic images of avatars for research use and model training.
Synthetic Data Validation: We demonstrate that our algorithm, trained exclusively on synthetic data performs on par with human raters for lateral shift, highlighting data synthesis as a viable solution for clinical domains with scarce data such as the domain of rare diseases.
Systematic Evaluation: We compare human and algorithmic performance on avatars and validate automated postural estimates against expert visual consensus on real patient images. We established a benchmark by having 20 experts rate 100 avatars and 100 patient images.
Clinical Application: We present an illustrative patient case to demonstrate the method’s potential for clinical decision support.

\section{Methods \& Materials} \label{methods}

\subsection{Automatic TWSTRS estimation}
\begin{figure}[htbp]
    \centering
    \includegraphics[width=1\linewidth]{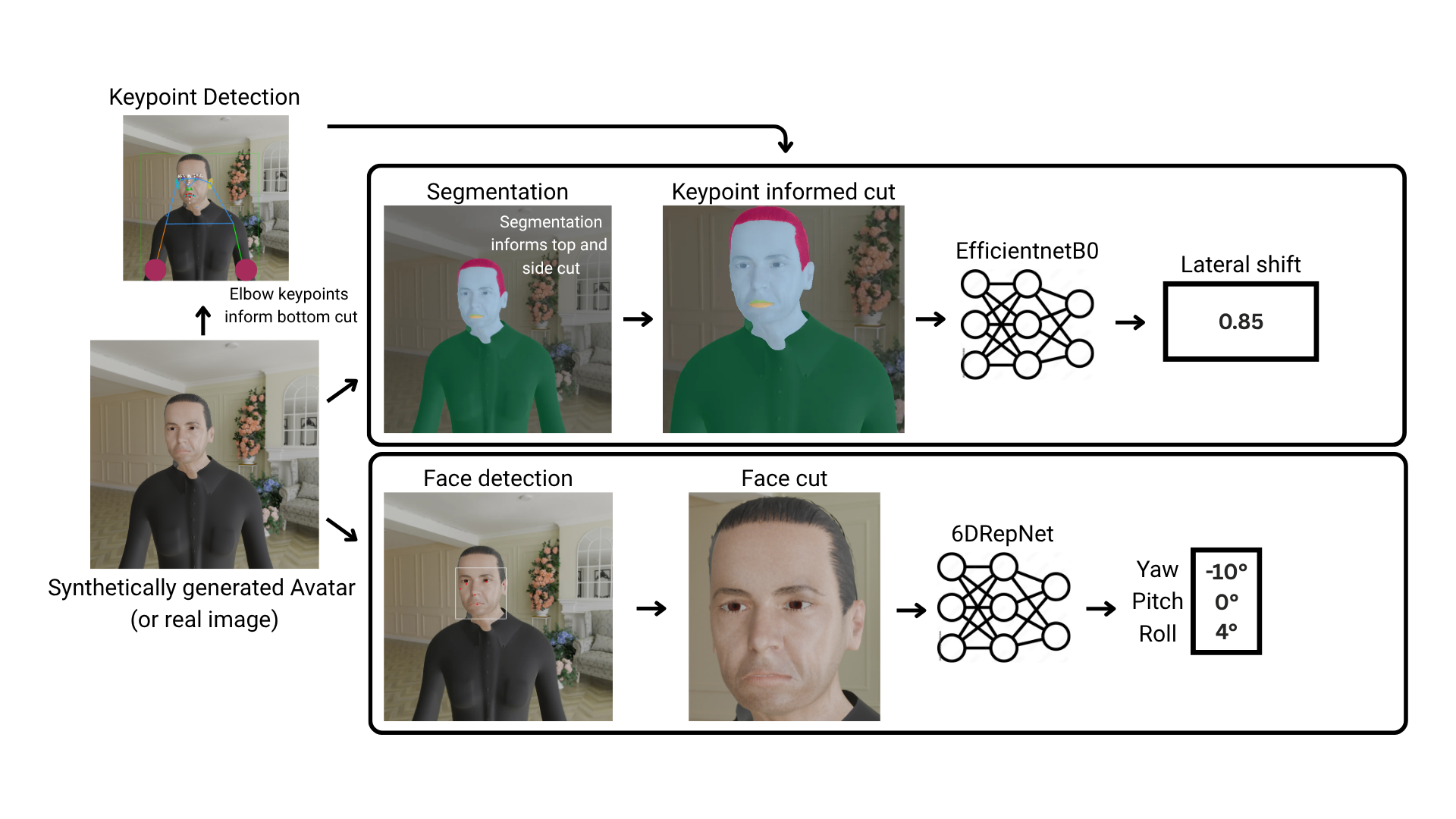}
    \caption{Computational pipeline for TWSTRS item estimation. TWSTRS=Toronto Western Spasmodic Torticollis Rating Scale.}
    \label{fig:pipeline}
\end{figure}

Two methods were used to estimate postural items: a pretrained head pose estimation neural network for rotational items and a neural network, pretrained exclusively on synthetic data for lateral shift. The combined pipeline is visualized in Fig.~\ref{fig:pipeline}.

\subsubsection{Head pose estimation for rotational TWSTRS items}
For head pose estimation, the pre-trained head pose estimation algorithm 6DRepNet by Hempel et al. \cite{hempel6dRotationRepresentation2022} is used on the frontal images. Prior to head pose estimation, facial regions are localized using the MTCNN face detector, and the image is cropped accordingly to ensure that the subsequent head pose estimation algorithm focuses specifically on the relevant region of the image. The head pose estimation algorithm follows a deep learning approach and regresses head orientation with a continuous 6D rotation representation, which is finally converted into Euler angles. These Euler angles are directly mappable to the corresponding TWSTRS item (yaw/torticollis, roll/laterocollis, pitch/(anterocollis/retrocollis)).
As the algorithm estimates continuous angle estimations and rarely predicts an exact zero, only predicted deviations above 5° were considered indicative of existence of the syndrome. This threshold minimizes the risk of misclassifying deviations unlikely to be clinically meaningful, i.e., avoiding classification of negligible rotations below 5° as abnormal (Supplementary Tab. \ref{tab:suppl1}).

\subsubsection{Avatar based pre-training for lateral shift}
For lateral shift estimation, following preprocessing steps were applied on the images: First, keypoint detection using Meta’s Sapiens model \cite{SapiensFoundationHuman} was performed. The (left and right) elbow keypoints were extracted, and the image cropped from the bottom at the height of the higher elbow keypoint. The top of the image was cropped based on the semantic segmentation (via Meta’s Sapiens as well) by removing all pixels above the uppermost boundary of the segmented regions. Next, the mass center of the head-neck region (as defined by the segmentation mask) was calculated. The images were then symmetrically expanded from this center point to the left and right until it reached a 4:3 aspect ratio.
The semantic segmentation generates a discrete mask per image that encodes 6 individual regions (head/neck, hair, clothes, skin, upper lip, and lower lip). The masks served as input to train EfficientNet‑B0 models \cite{tanEfficientNetRethinkingModel2019}. Images were resized to 224×224 px to fit the optimal input size of EfficientNet-B0. 
The models were trained using mean squared error loss, optimized with the AdamW algorithm (initial learning rate was $5\times 10^{-4}$). Training includes various data augmentations to prevent overfitting on the training dataset: random perspective warping, stretching or compression with padding, horizontal flipping, and random image cropping (see Supplementary Tab. \ref{tab:suppl2} \& \ref{tab:suppl3} for detailed augmentation and training parameter information).
For evaluation, we determined the threshold for mapping the continuous regression output to a binary lateral-shift label by selecting the value that maximized TPR + accuracy on the synthetic validation set.

\subsection{Datasets}
\subsubsection{Synthetic dataset for algorithmic training}
\begin{figure}[htbp]
    \centering
    \includegraphics[width=1\linewidth]{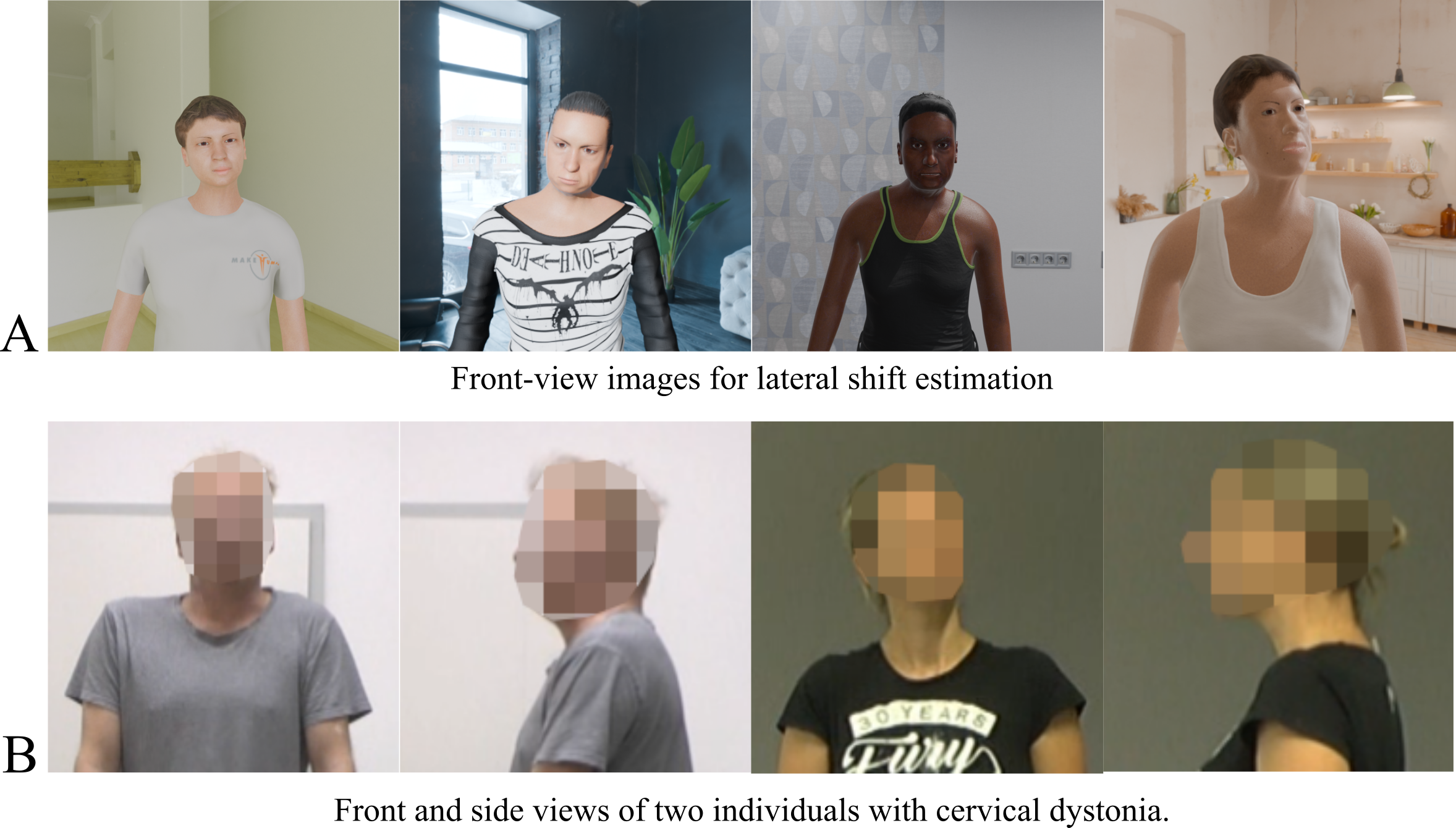}
    \caption{Synthetic avatar dataset and clinical examples of cervical dystonia. (A) Examples of synthetic avatar images used for model training. The dataset includes frontal renderings with varying camera perspectives, lighting, backgrounds, clothing, and morphologies across a range of symptom severities. (B) Front and side views of two individuals with cervical dystonia. Left: Characterized by mild torticollis (0.5), laterocollis (0.5), and a prominent lateral shift (0.7). Right: A complex syndrome featuring torticollis (2.5), laterocollis (0.9), and minimal retrocollis (0.5). Reported values represent the average of 10 independent clinical ratings using static items of the Toronto Western Spasmodic Torticollis Rating Scale (TWSTRS).}
    \label{fig:avatars}
\end{figure}

We generated a large‑scale synthetic dataset of avatars showing symptoms of CD (Fig.~\ref{fig:avatars}A). We rendered about 16,000 frontal‑view avatars with head and neck translated in opposite directions across a normalized range 0–1 (0=no shift, 1=12.5° maximal opposing rotation of head and neck), thereby showing a lateral shift.
Additional cervical rotations were introduced to more fully capture the heterogeneity of real-world symptoms, as a lateral shift rarely occurs in isolation. Six cervical degrees of freedom, i.e., yaw, pitch, and roll for both head and neck rotations, as conceptualized by the Col-Cap concept \cite{finstererCollumcaputCOLCAPConcept2015}, were independently perturbed using stochastic sampling. In 80\% of the cases, individual angles were sampled using Gaussian distribution ($\mu$=0, $\sigma$=10°); in the remaining 20\%, angles were held at zero.

The open-source rendering software blender \cite{foundationCommunity} is used for avatar creation, along with the MPFB add-on \cite{MPFB} for human model creation. The add-on allows for detailed customization of human morphology via parametric sliders (e.g., age, gender, body proportions) and automatically generates a rigged, textured mesh suitable for rendering. For the datasets, all parameters are sampled script-based and randomized. Furthermore, a variety of randomly chosen assets for clothes from the MPFB add-on, and environment textures from Poly Haven \cite{havenPolyHavenPoly} are included.

\subsubsection{Patient image dataset for algorithm evaluation}
\begin{figure}[htbp]
    \centering
    \includegraphics[width=1\linewidth]{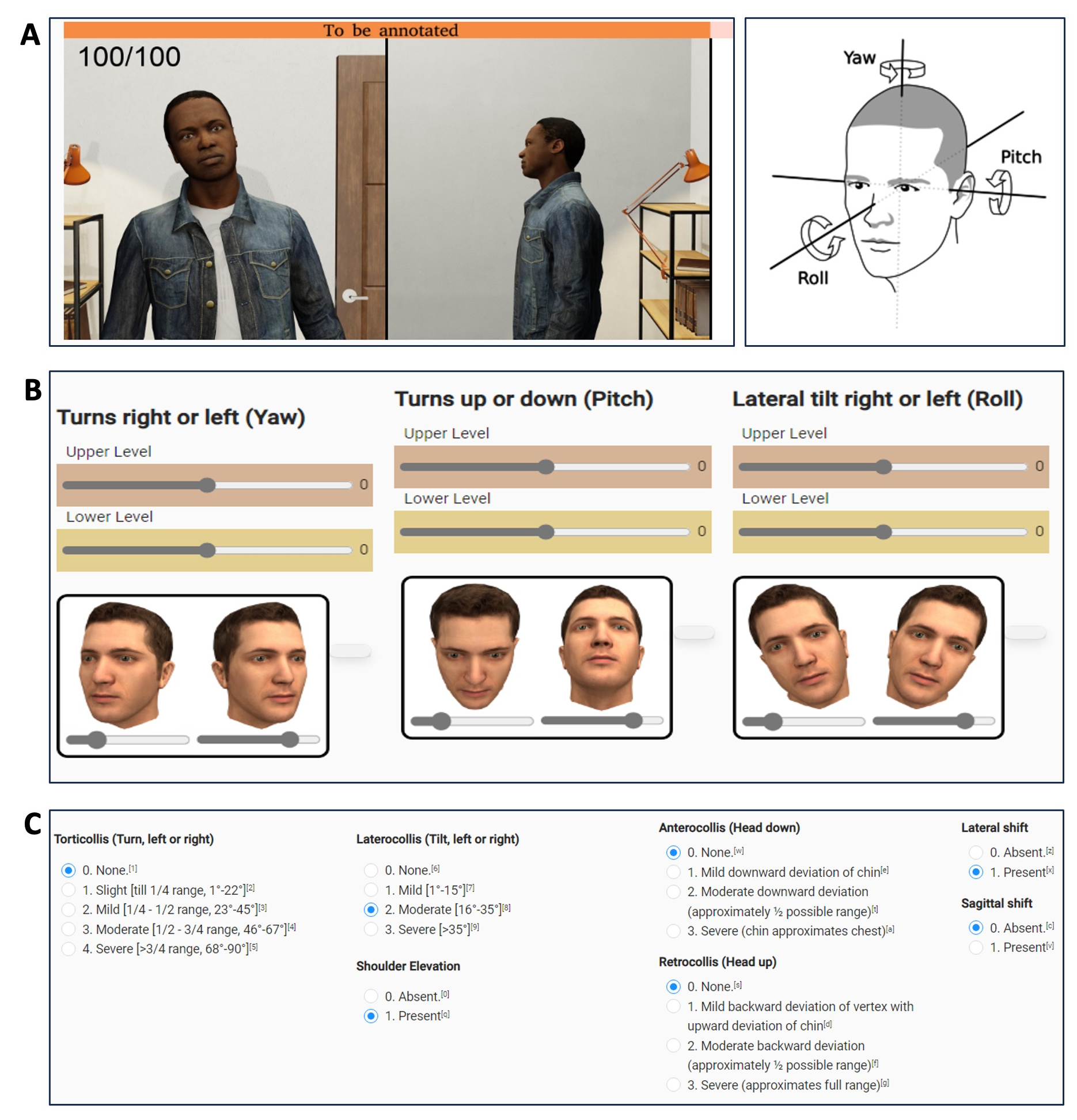}
    \caption{The annotation platform; A synthesized avatar demonstrating compound head and neck movements along the three principal axes of rotation: yaw (horizontal turning), pitch (vertical nodding), and roll (side-to-side tilting). (B) The Col-Cap evaluation interface, which uses interactive sliders to separately rate upper and lower cervical motion in each rotational plane. (C) Clinical scoring using the TWSTRS motor subscale, with categorical ratings for cervical postures including torticollis, laterocollis, anterocollis, retrocollis, shoulder elevation, and sagittal displacement. In this study, only TWSTRS estimations were used.}
    \label{fig:suppl1}
\end{figure}

A dataset consisting of 100 avatar images and 100 patient pictures, each from the front and side perspective was created and evaluated by 20 clinical raters in a multi-center study (Supplementary Fig. \ref{fig:suppl1}). Each rater assessed a subset of 50 avatar- and 50 real patient- images, ensuring that each image received approximately 10 independent ratings. Each image was rated on static TWSTRS items, including rotational (torticollis, laterocollis, anterocollis/retrocollis) and the lateral shift item \cite{loensValidationClinicalRatings2026}. We assessed the mean rating, which is the arithmetic average of the individual expert ratings per image. This averaged score served as a "super-annotation”, effectively reducing random rating errors and providing a more stable and continuous approximation of symptom severity. Since TWSTRS ratings are based on discrete ordinal categories, small differences in symptom expression may not be captured by individual ratings alone. Our aggregation enabled finer gradations of symptom severity assessment for detecting subtle signs, such as minor shifts. Fig.~\ref{fig:avatars}B shows two patient examples. The left subject shows a mild rotational deviation with a lateral shift, while the right subject shows a mix of rotational and translational components.

\subsection{Evaluation}
For clinical evaluation, we assessed our algorithm on a held-out subset of synthetic avatars not used during training. For such, we used a different synthetic dataset generated with Rocketbox avatars in Blender and a constant background, independent of the PolyHaven/MPFG pipeline used for training. This reduces the risk of model overfitting within the synthetic-data regime.
Evaluation on the avatars dataset enabled a direct comparison between the model’s predictions and the ground-truth TWSTRS items, and allowed us to evaluate how its performance relates to that of human raters who annotated the same avatar image test set. Furthermore, to confirm that our approach generalizes across domains and is applicable to real-world settings, we evaluated it on 100 real patient photographs, with 20 clinical experts each annotating 50 images so that every photograph was rated by 10 independent raters.
Performance was evaluated in two ways: Correlation with expert ratings on real patient images and benchmarking against ground-truth values on synthetic avatars.
For the clinical evaluation, we used mean rating per image as reference ratings as defined in 2.2.2. The mean rating allows finer-grained assessment and provides a continuous approximation of symptom severity.
Pearson’s correlation coefficient (r) was calculated to assess the linear association between the algorithm’s estimates and the mean expert ratings for each TWSTRS item. Correlation values were interpreted in the context of the score distributions.
For evaluation on avatars with predefined head and neck deviations, evaluation against the known ground-truth values was performed via classification accuracy and true positive rate (TPR) for each TWSTRS item individually. Results were compared to both the individual human ratings and the mean ratings, allowing for an objective assessment of performance.
To quantify agreement between annotators, we calculated inter-rater reliability for all TWSTRS items using Krippendorff’s alpha, using a nominal metric for lateral shift, and ordinal metric for rotational items. These values are used to contextualize the correlation between human annotations and algorithmic estimates.

\section{Results} \label{results}
We report the performance of the proposed algorithm for automated head-pose and lateral shift estimation, evaluating its correlation with corresponding postural TWSTRS items across two datasets: 100 real patient photographs, each annotated by up to 10 clinical experts and 100 synthetic avatar images with known head and neck rotations. To verify that the synthetic avatars serve as a valid proxy for real patients, we additionally compared inter-rater reliability (IRR) between human ratings on avatar and real images.

\subsection{Confidence Interval Comparison of Human Estimates on Avatars vs. Real Images}
Human ratings on synthetic avatars and real patient photographs showed similar inter-rater reliability across most TWSTRS items. Tab. 1 summarizes Krippendorff’s α with 95\% bootstrap confidence intervals for both datasets. Confidence intervals were estimated via nonparametric bootstrapping, resampling images with replacement and recalculating Krippendorff’s α across 2,000 iterations. For lateral shift, torticollis, and laterocollis, the confidence intervals for avatar and real images overlapped, indicating no detectable difference in rater consistency. For anterocollis/retrocollis, the confidence intervals did not overlap, with avatars showing higher reliability. Because raters showed similar levels of agreement, we argue that the generated avatars appear to mirror the visual ambiguity and distinctness of real-world symptoms, supporting their validity as a dataset for evaluation.

\begin{table}[h]
\centering
\caption{Inter-rater reliability (Krippendorff’s $\alpha$) of human ratings with 95\% bootstrap confidence intervals for synthetic avatar and real patient images.}
\label{tab:reliability}
\begin{tabular}{@{}llll@{}}
\toprule
TWSTRS item & Avatars ($\alpha$ [CI]) & Real images ($\alpha$ [CI]) & CI Overlap \\ \midrule
Torticollis & 0.67 [0.58-0.74] & 0.64 [0.53-0.72] & Yes \\
Laterocollis & 0.47 [0.37-0.56] & 0.48 [0.39-0.56] & Yes \\
\begin{tabular}[c]{@{}l@{}}Anterocollis/\\ Retrocollis\end{tabular} & 0.81 [0.73-0.86] & 0.56 [0.46-0.64] & No (avatars better) \\
Lateral shift & 0.27 [0.19-0.35] & 0.31 [0.20-0.40] & Yes \\ \bottomrule
\end{tabular}
\end{table}

\subsection{Comparison on avatars: human rating vs. algorithm}
\begin{figure}[htbp]
    \centering
    \includegraphics[width=1\linewidth]{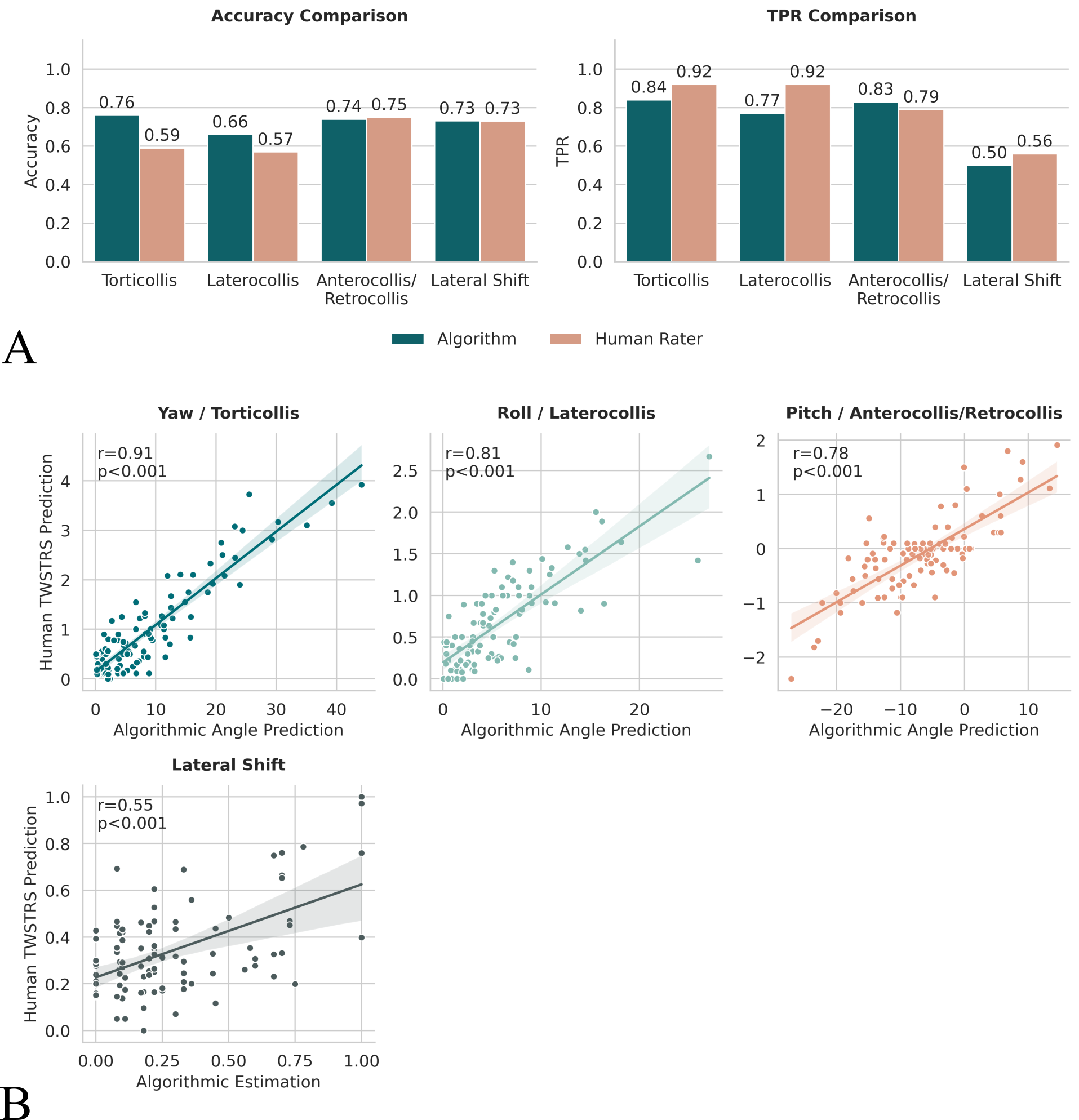}
    \caption{Performance and clinical correlation of the TWSTRS algorithm. (A) Comparison of algorithm performance (true positive rate and accuracy) against human raters on the avatar benchmark with known ground-truth head deviations. (B) Correlation between TWSTRS estimations and mean clinical ratings on real-world clinical images across five items: torticollis, laterocollis, anterocollis/retrocollis, and lateral shift. Each dot represents a single patient image. TWSTRS = Toronto Western Spasmodic Torticollis Rating Scale.}
    \label{fig:performance}
\end{figure}

In the evaluation on avatar data, the algorithm's head-pose and lateral shift estimations were compared to direct TWSTRS item ratings from human experts, using the predefined angular rotations as an objective ground truth. Fig.~\ref{fig:performance}A reports two metrics: The true positive rate (TPR), which measures how often positive cases are correctly identified, and overall accuracy, which measures the proportion of all correct classifications. Two types of estimates are shown: (1) the mean of all human annotations for each task and (2) the algorithm’s estimations. When compared to the mean of human ratings, the algorithm achieved higher TPR for anterocollis/retrocollis, while humans performed better for torticollis, laterocollis, and lateral shift. In terms of accuracy, the algorithm outperforms human ratings for all items, except for anterocollis/retrocollis, and is on par with lateral shift. The algorithm’s lower TPR but bigger accuracy in some items suggests greater specificity (see Supplementary Tab. \ref{tab:suppl4} for exact values).

\subsection{Human ratings vs. algorithm on real images}
First, we assessed how well the algorithm correlates with human ratings on real-world clinical images to address its potential for clinical applicability. Fig.~\ref{fig:performance}B shows the Pearson correlation between the algorithm’s estimations and expert human ratings for TWSTRS items on 100 real patient images, each annotated by 10 clinical experts on average. The results indicate high agreement for rotational symptoms, with large correlations for torticollis (r=0.91), laterocollis (r=0.81), and anterocollis/retrocollis (r=0.78). The algorithm showed medium correlation for lateral shift (r=0.55). All results are highly significant with p<0.001.

\section{Discussion} \label{discussion}
We present an algorithmic pipeline that shows a strong correlation with clinical TWSTRS estimates for rotational items. In the example of lateral shift, we demonstrated that the generation of synthetic data is a valid approach to overcome data scarcity and to enable data-driven model training for symptoms that are rare and difficult to annotate consistently. Our synthesis approach compensates by producing rare, but clinically plausible, and systematically varied characteristics of the lateral shift, which are often underrepresented in clinical data.
On our synthetic-image benchmark, the algorithm outperformed the average accuracy of ten clinical raters for rotational items and achieved comparable performance for lateral shift. The evaluation on avatars revealed a difference between the algorithm and human estimates. Human raters were more sensitive, meaning they rarely missed a symptom when it was present. However, this high sensitivity came at the cost of overall accuracy. This suggests that humans tend to 'over-diagnose' by identifying symptoms even in healthy avatars. The algorithm was stricter: while it missed some mild cases, it was right more often overall because it produced fewer false positives. This situation was only flipped for anterocollis/retrocollis, where the algorithm was more sensitive. Finally, for the lateral shift, the algorithm performed on par with humans in accuracy, confirming that training with synthetic images is a working strategy.
The lateral shift was notably harder to estimate correctly than rotational items, which may be due to the greater complexity of the underlying task. While rotational scores can be derived from head orientation alone, shift estimation relies on visual indications related to both head position and the trunk mid-line, with respect to their spatial offset. This introduces additional sources of error, which may explain the lower model performance observed for lateral shift. Importantly, the moderate correlation with expert ratings on patient images (r=0.55) should be interpreted in the context of limited inter-rater reliability for this item (as reported in Table 1), meaning that algorithmic and human uncertainties compound in correlation-based evaluations. When the model is evaluated against error-free ground truth in the synthetic avatar benchmark, it performs at a level comparable to clinicians. This suggests that its lower correlation on real images reflects the difficulty and inconsistency of the human reference, rather than exclusively limitations of the algorithm itself.
Existing vision-based approaches addressed only rotational items \cite{yePilotFeasibilityStudy2022,peachHeadMovementDynamics2024,zhangHoldThatPose2022} or included shift items with a limited sample size \cite{nakamuraPilotFeasibilityStudy2019}. In contrast, our approach achieves the highest correlations across all rotational items and further is the first example of a lateral shift estimator with significant correlation on human ratings, which expands the scope of automated assessment (Supplementary Tab. \ref{tab:suppl5}. 
However, a primary limitation is that the algorithm uses only static visual data to quantify head pose, which differs from the clinical assessment of CD. Assessment includes non-visual features like range of motion, sensory tricks, and postural variability during tasks. Consequently, our system acts as an objective visual proxy for specific dystonic movements. It is not a replacement for a comprehensive examination. While rater reliability was comparable between synthetic avatars and real patients, the geometric ground truth of an avatar does not capture all anatomical nuances. Clinicians must differentiate between movements at the upper cervical levels (caput) and the lower cervical spine (collis) to select muscles for injection.
Our framework operates exclusively on static image-frames and does not evaluate temporal dynamics. Nevertheless, the underlying approach is well-suited for extension to temporal analysis. Its architecture allows for image-frame wise inference over videos, enabling the capture of dynamic symptom expression once integrated with structured, time-aligned inputs: We proposed Move2Screen \cite{schulzeUsabilityAdherenceEvaluation2025, stengerAndroidAppSymptomatic2025} as a protocol-guided application that allows patients to record standardized video assessments in the home environment. With the use of a temporally consistent video protocol (Supplementary Tab. \ref{tab:suppl6}), the app guides the patient through specific head movements (e.g., neutral midline, rotations) to reveal dystonic symptoms. However, high-frequency video collection creates a logistical bottleneck: manual rating of weekly videos for a single patient cohort is practically impossible due to the sheer volume of data. The protocol-driven structure ensures that posture and movement are assessed in a consistent and temporally coherent manner across videos, making the resulting video sequences particularly suitable for automated analysis.

\begin{figure}[htbp]
    \centering
    \includegraphics[width=1\linewidth]{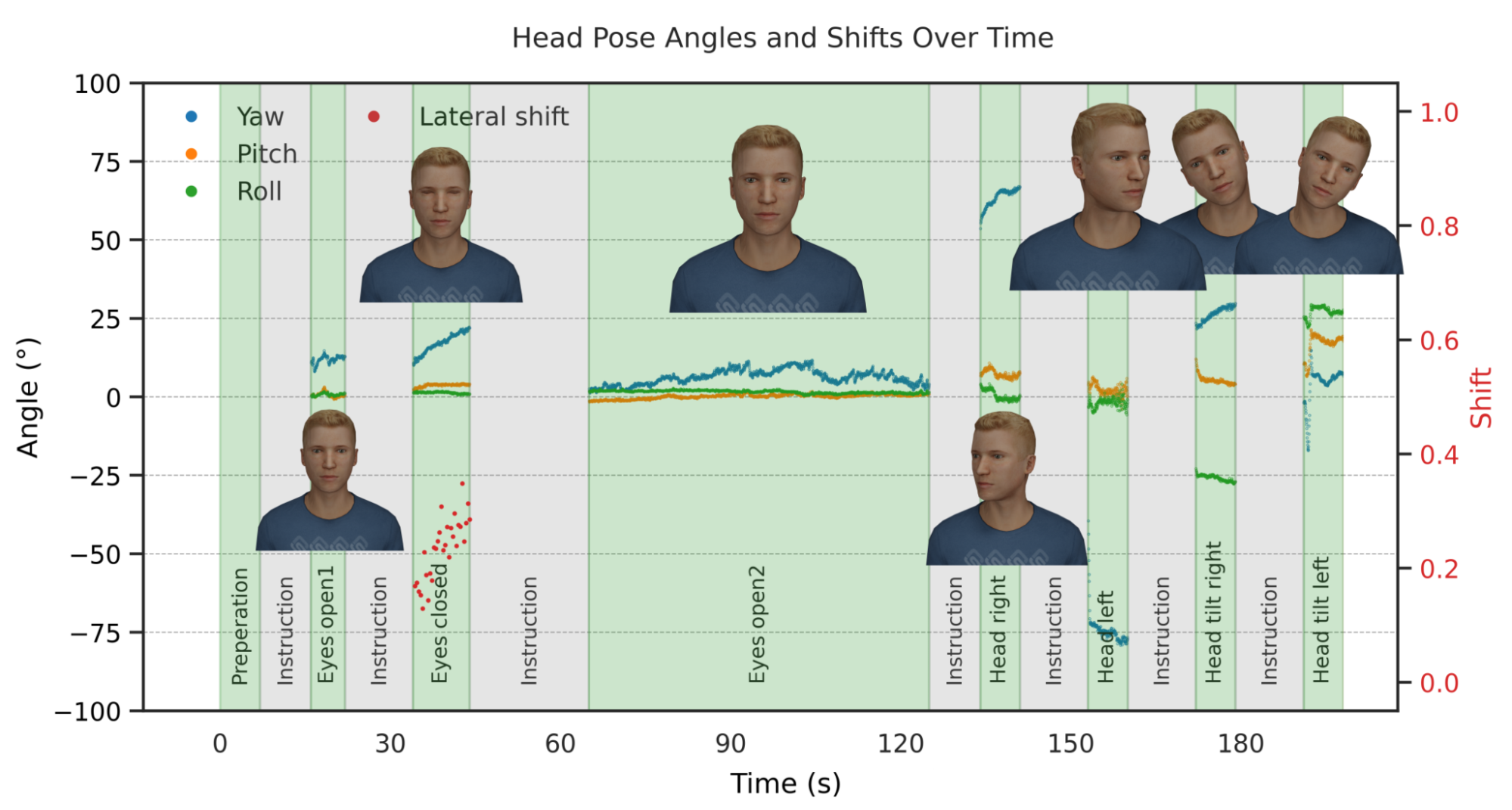}
    \caption{Predictions of our proposed algorithm on each frame of a video, recorded by the Move2Screen app as an illustrative example use case of the algorithm. The tasks (green areas) are chosen to make symptoms accessible, like the maximal possible rotation, or the task to let the head move into the most comfortable position.}
    \label{fig:predictions}
\end{figure}

Fig.~\ref{fig:predictions} illustrates how this integration can work in practice. It shows frame-by-frame head pose and lateral shift predictions generated from a video recorded using the Move2Screen protocol (Supplementary Tab. \ref{tab:suppl6}). Green-shaded intervals indicate specific tasks, time-consistent between all videos recorded by the app. During the Eyes-closed task, where patients are asked to let their head fall into its most relaxed position, the model detects a progressive yaw drift toward the patient’s primary dystonic direction, consistent with the actual clinical diagnosis. A similar head position is seen in the subsequent Eyes-open-2 task, where the patient tries to face the camera but shows ongoing deviation. During the Head-right and Head-left tasks, the model shows a smaller range of motion on one side compared to the other. These task-specific differences demonstrate how protocol-guided video analysis can capture both static postures and movement limitations, supporting more complete and structured assessment of symptom severity.
With our machine-learning approach for estimation, we envision together with the Move2Screen app a scalable system for high-frequency, home-based monitoring of cervical dystonia. While the app enables standardized video recordings, the algorithm provides automated analysis of data that would be costly and time consuming to evaluate manually. In this way, the Move2Screen framework offers a foundation for telemedical monitoring in both clinical care and research. The clinical ability to capture relevant changes in head positioning and rotational TWSTRS items in patients with CD needs to be proven in real-world patient settings.Finally, this work demonstrates the utility of synthetic data in training AI systems for underrepresented clinical conditions. Confirming that synthetic augmentation can overcome data scarcity and benchmarking challenges, our work may serve as a blueprint for developing similar tools in other domains of neurology and beyond.

\subsection{Acknowledgment}
This preprint was created using the LaPreprint template (\url{https://github.com/roaldarbol/lapreprint}) by Mikkel Roald-Arb\o l \textsuperscript{\orcidlink{0000-0002-9998-0058}}.

\subsection{Author contributions}
Conceptualization: R.S., S.L., S.F., and T.B., Data curation: R.S., S.L., F.H., A.M., T.P., A.W., G.M.S., T.U., M.B., M.G.P., L.M.L., M.A.H., R.H., A.L.A.M., T.M., F.S., L.L.G., J.R., K.E.Z., J.H., J.B., A.B., R.W., C.W.I., T.O., D.Z., L.H.R., J.I.L., P.A., P.N., J.R., J.U., K.K., J.D.L., J.H., L.V.S., S.Z., K.G., P.T., A.L., T.B., S.F., Formal analysis: R.S., Funding acquisition: S.F. and T.B., Investigation: R.S., Methodology: R.S., S.L., N.B., S.F., and T.B., Project administration: R.S., S.L., S.F., and T.B., Resources: R.S., S.L., S.F., and T.B., Software: R.S., Supervision: S.F. and T.B., Validation: R.S., Visualization: R.S., S.L., Writing – original draft: R.S., Writing – review \& editing: R.S., S.L., N.B., S.F., and T.B.

\subsection{Supplementary}

\begin{table}[h]
\centering
\caption{TWSTRS scores from angle ranges}
\vspace{2mm}
\begin{tabular}{clll}
\toprule
\textbf{Score} & \textbf{Torticollis} & \textbf{Laterocollis} & \textbf{Anterocollis/Retrocollis} \\ \midrule
0 & Absent (0°–4°) & Absent (0°–4°) & Absent (0°–4°) \\
1 & Slight (5°–22°) & Mild (5°–15°) & Mild (5°–15°) \\
2 & Mild (23°–45°) & Moderate (16°–35°) & Moderate (16°–35°) \\
3 & Moderate (46°–67°) & Severe ($>$35°) & Severe ($>$35°)* \\
4 & Severe (68°–90°) & - & - \\ \bottomrule
\end{tabular}
\label{tab:suppl1}
\end{table}

\begin{table}[h]
\centering
\caption{Data augmentation during training time.}
\vspace{2mm}
\begin{tabularx}{\textwidth}{@{}Xlll@{}}
\toprule
\textbf{Transformation} & \textbf{Key parameter} & \textbf{Probability} \\ \midrule
Horizontal Flip & - & $p=0.5$ \\ \addlinespace
Random stretch and compress & scale range = [0.7, 1.2] & $p=0.4$ \\ \addlinespace
Random cutout & \begin{tabular}[c]{@{}l@{}}min fraction = 0.1, \\ max fraction = 0.4\end{tabular} & $p=0.2$ \\ \addlinespace
Random rotation & degree range: [$-10^\circ$, $+10^\circ$] & $p=0.3$ \\ \addlinespace
Random resized crop & \begin{tabular}[c]{@{}l@{}}scale range = [0.6, 1.0], \\ size = 240$\times$240\end{tabular} & $p=0.6$ \\ \bottomrule
\end{tabularx}
\label{tab:suppl2}
\end{table}

\begin{table}[h]
\centering
\caption{Model, training, and optimization hyper-parameters.}
\vspace{2mm}
\begin{tabular}{@{}ll@{}}
\toprule
\textbf{Component} & \textbf{Setting} \\ \midrule
Backbone & EfficientNet‑B0, $1 \times 224 \times 224$ input \\
Regression head & Fully-connected (FC) $\to$ lateral shift \\
Loss function & \begin{tabular}[c]{@{}l@{}}Correlation aware loss: $0.7 \times \text{MSE-loss} + 0.3 \times \text{Concordance Corr. Coef.}$\\ on training set; L1-norm on validation set\end{tabular} \\
Optimizer & AdamW (lr=$5 \times 10^{-4}$, weight\_decay=$1 \times 10^{-4}$) \\
Scheduler & \begin{tabular}[c]{@{}l@{}}ReduceLROnPlateau (monitor=val\_corr, factor=0.5, patience=10, \\ min\_lr=$1 \times 10^{-6}$)\end{tabular} \\
Batch size & 32 \\
Train/Validation-Split & 90\%/10\% \\ \bottomrule
\end{tabular}
\label{tab:suppl3}
\end{table}

\begin{table}[h]
\centering
\caption{Evaluation of human and algorithmic ratings.}
\vspace{2mm}
\begin{tabular}{@{}llll@{}}
\toprule
\textbf{Metric} & \textbf{Category} & \textbf{Human rating} & \textbf{Algorithm} \\ \midrule
TPR & Torticollis & 92\% & 84\% \\
 & Laterocollis & 92\% & 77\% \\
 & Anterocollis/Retrocollis & 79\% & 83\% \\
 & Lateral Shift & 56\% & 50\% \\ \addlinespace
Accuracy & Torticollis & 59\% & 76\% \\
 & Laterocollis & 57\% & 66\% \\
 & Anterocollis/Retrocollis & 75\% & 74\% \\
 & Lateral Shift & 73\% & 73\% \\ \bottomrule
\end{tabular}
\label{tab:suppl4}
\end{table}

\begin{table}[h]
\centering
\caption{Reported correlations between algorithmic estimates and human ratings on TWSTRS items.}
\vspace{2mm}
\begin{tabular}{lccccc}
\toprule
\textbf{Study} & \textbf{Torticollis} & \textbf{Laterocollis} & \textbf{\begin{tabular}[c]{@{}c@{}}Anterocollis/\\ Retrocollis\end{tabular}} & \textbf{Lateral Shift} & \textbf{Sagittal Shift} \\ \midrule
Nakamura et al. & 0.902 & 0.369 & 0.181 & $-0.50$ & 0.557 \\
Ye et al. & 0.843 & 0.667 & 0.701 & --- & --- \\
Zhang et al. & 0.62 & 0.60 & 0.59 & --- & --- \\
Peach et al. & 0.79 & 0.67 & 0.66 & --- & --- \\
Our & 0.91 & 0.81 & 0.78 & 0.55 & --- \\ \bottomrule
\end{tabular}
\label{tab:suppl5}
\end{table}

\begin{table}[h]
\centering
\caption{Video-protocol as implemented in Move2Screen.}
\vspace{2mm}
\begin{tabularx}{\textwidth}{@{}Xl@{}}
\toprule
\textbf{Task} & \textbf{Time} \\ \midrule
Preparation & 7 sec \\
Instruction & 9 sec \\
With eyes open, looking at the camera & 6 sec \\
Instruction & 12 sec \\
Eyes closed while the subject moves their head in the most comfortable direction, revealing involuntary dystonic tendencies without visual correction. & 10 sec \\
Instruction & 21 sec \\
Subject holds head at the neutral midline to assess posture maintenance and onset of dystonia. & 60 sec \\
Instruction & 9 sec \\
Subject turns head to the right to assess horizontal rotational range. & 7 sec \\
Instruction & 12 sec \\
Subject turns head to the left to assess horizontal rotational range. & 7 sec \\
Instruction & 12 sec \\
Subject tilts head to the right to evaluate lateral tilt capabilities. & 7 sec \\
Instruction & 12 sec \\
Subject tilts head to the left to evaluate lateral tilt capabilities. & 7 sec \\
Position Change: The subject rotates the chair by 90° to enable a side view & 20 sec \\
Instruction & 8 sec \\
Subject sits sideways and holds the head straight to observe sagittal abnormalities like antero- or retrocollis. & 10 sec \\
Instruction & 9 sec \\
Subject lifts head upward to assess neck extension from the profile view. & 7 sec \\
Instruction & 12 sec \\
Subject lowers head downward to assess neck flexion from the profile view. & 7 sec \\ \bottomrule
\end{tabularx}
\label{tab:suppl6}
\end{table}

\subsubsection{TWSTRS-mapping from continuous angles}
A design choice is that the model relies on a hard-coded threshold of 5° to identify pathological rotation when mapping continuous angle estimations to TWSTRS categories. This choice reflects a practical trade-off: Without such a threshold, the model would most often produce a score of at least 1 even for minimal, likely non-pathological deviations. Although TWSTRS defines score 1 as any deviation greater than 0°, this threshold is not interpreted literally in practice. Rather, it serves as a guideline for clinicians to detect a clearly visible rotation, not to quantify imperceptible movements. Thus, the 5° cutoff provides a clinically meaningful lower bound that may reflect how experts tend to score mild presentations.

\subsubsection{Mismatch between human and algorithmic estimations}
Although the correlation between algorithmic and expert ratings is high for rotational TWSTRS items, a closer examination reveals systematic mismatches in absolute angular values. For example, while human raters often assign a retrocollis close to 2 or 2 exactly, corresponding to an angular range of 16–36°, the algorithm rarely predicts more than 15° for these same cases. A similar pattern is observed in torticollis, where expert ratings frequently reach level 4, but algorithmic estimates do not exceed 45°, which lies at the upper limit of a TWSTRS score of 2. In contrast, anterocollis predictions align more closely with human ratings. For laterocollis, the discrepancy is less pronounced but still present, with the algorithm tending to predict slightly lower angles than those implied by expert scores. These discrepancies may come from both overestimation by human raters and underestimation by the algorithm. In previous work, we have shown that clinicians tend to systematically overrate rotational items on the TWSTRS scale 1. On the algorithmic side, one likely source of underestimation is the use of head pose estimation relative to the camera. However, some patients show compensatory trunk or body rotations, the head may appear less rotated in camera space, leading to reduced angular predictions.
\clearpage
\printbibliography

\if@endfloat\clearpage\processdelayedfloats\clearpage\fi






\end{document}